%% file: root.tex
\documentclass[letterpaper, 10 pt, journal, twoside]{IEEEtran} 

\usepackage{amsmath,amsfonts}
\usepackage{algorithmic}
\usepackage{algorithm}
\usepackage{array}
\usepackage[caption=false,font=normalsize,labelfont=sf,textfont=sf]{subfig}
\usepackage{textcomp}
\usepackage{stfloats}
\usepackage[table]{xcolor}
\usepackage{colortbl}
\usepackage{url}
\usepackage{verbatim}
\usepackage{graphicx}
\usepackage{cite}
\usepackage{multirow}
\usepackage{booktabs}
\usepackage{pifont}
\usepackage{enumitem}
\usepackage{makecell}
\usepackage{colortbl}
\usepackage{tcolorbox}
\usepackage{float}
\usepackage{placeins}
\usepackage{xcolor}
\definecolor{lightgreen}{rgb}{0.56, 0.93, 0.56} % Define lightgreen color
\usepackage{listings}
\usepackage{footmisc} % To control footnotes
\usepackage{caption}
\usepackage{colortbl} 
\usepackage{xcolor}
\usepackage{hyperref}
\hypersetup{
    colorlinks=false, % disables colored links
    pdfborder={0 0 0}, % removes borders around links
}

\definecolor{blue}{RGB}{0,0,255}
\definecolor{green}{RGB}{81,157,69}
\definecolor{red}{RGB}{255,0,0}

\begin{document}

\title{Embodied-RAG: General Non-parametric Embodied Memory for Retrieval and Generation}

% \title{General non-parametric Embodied Memory}

% The \author macro works with any number of authors. There are two
% commands used to separate the names and addresses of multiple
% authors: \And and \AND.
%
% Using \And between authors leaves it to LaTeX to determine where to
% break the lines. Using \AND forces a line break at that point. So,
% if LaTeX puts 3 of 4 authors names on the first line, and the last
% on the second line, try using \AND instead of \And before the third
% author name.

% NOTE: authors will be visible only in the camera-ready and preprint versions (i.e., when using the option 'final' or 'preprint'). 
% 	For the initial submission the authors will be anonymized.

% \author{
%   Quanting Xie\textsuperscript{*} \hspace{2em} So Yeon Min\textsuperscript{*} \hspace{2em} Tianyi Zhang \hspace{2em} \\
%   Aarav Bajaj \hspace{2em} Ruslan Salakhutdinov \hspace{2em} Matthew Johnson-Roberson\hspace{2em} Yonatan Bisk\\
%   Carnegie Mellon University\\ 
%   \texttt{quantinx@andrew.cmu.edu} \hspace{2em} \texttt{soyeonm@andrew.cmu.edu}\\
% \markboth{IEEE Robotics and Automation Letters. Preprint Version. Accepted Month, Year}
% {Xie \MakeLowercase{\textit{et al.}}: Embodied-RAG}  

% Make room for more info lines in the \author command  
\author{Quanting Xie$^{1}$, So Yeon Min$^{1}$, Pengliang Ji$^{1}$, Yue Yang$^{2}$, Tianyi Zhang$^{1}$, Kedi Xu$^{1}$, Aarav Bajaj$^{1}$,  Ruslan Salakhutdinov$^{1}$, Matthew Johnson-Roberson$^{1}$, and Yonatan Bisk$^{1}$%
% \thanks{Manuscript received: Month, Day, Year; Revised Month, Day, Year; Accepted Month, Day, Year.}%Use only for final RAL version
\thanks{This material is based upon work supported by the Defense Advanced Research Projects Agency (DARPA) under Agreement No. HR00112490375 and partially supported by funding from Lockheed Martin Corporation.}%Use only for final RAL version
\thanks{$^{1}$Quanting Xie, So Yeon Min, Pengliang Ji, Tianyi Zhang, Kedi Xu, Aarav Bajaj, Ruslan Salakhutdinov, Matthew Johnson-Roberson, and Yonatan Bisk are with Carnegie Mellon University, Pittsburgh, PA 15213, USA}
\thanks{$^{2}$Yue Yang with Apochs, Inc, CA 94560, USA}
\thanks{Corresponding author: {\tt\footnotesize  quantinx@andrew.cmu.edu}}
}
%

% \thanks{Digital Object Identifier (DOI): see top of this page.}
  %% examples of more authors
  %% \And
  %% Coauthor \\
  %% Affiliation \\
  %% Address \\
  %% \texttt{email} \\
  %% \AND
  %% Coauthor \\
  %% Affiliation \\
  %% Address \\
  %% \texttt{email} \\
  %% \And
  %% Coauthor \\
  %% Affiliation \\
  %% Address \\
  %% \texttt{email} \\
  %% \And
  %% Coauthor \\
  %% Affiliation \\
  %% Address \\
  %% \texttt{email} \\
% }

\maketitle

%===============================================================================

\begin{figure*}[!ht] % Use figure* to span two columns
    \centering
    \includegraphics[width=1\textwidth]{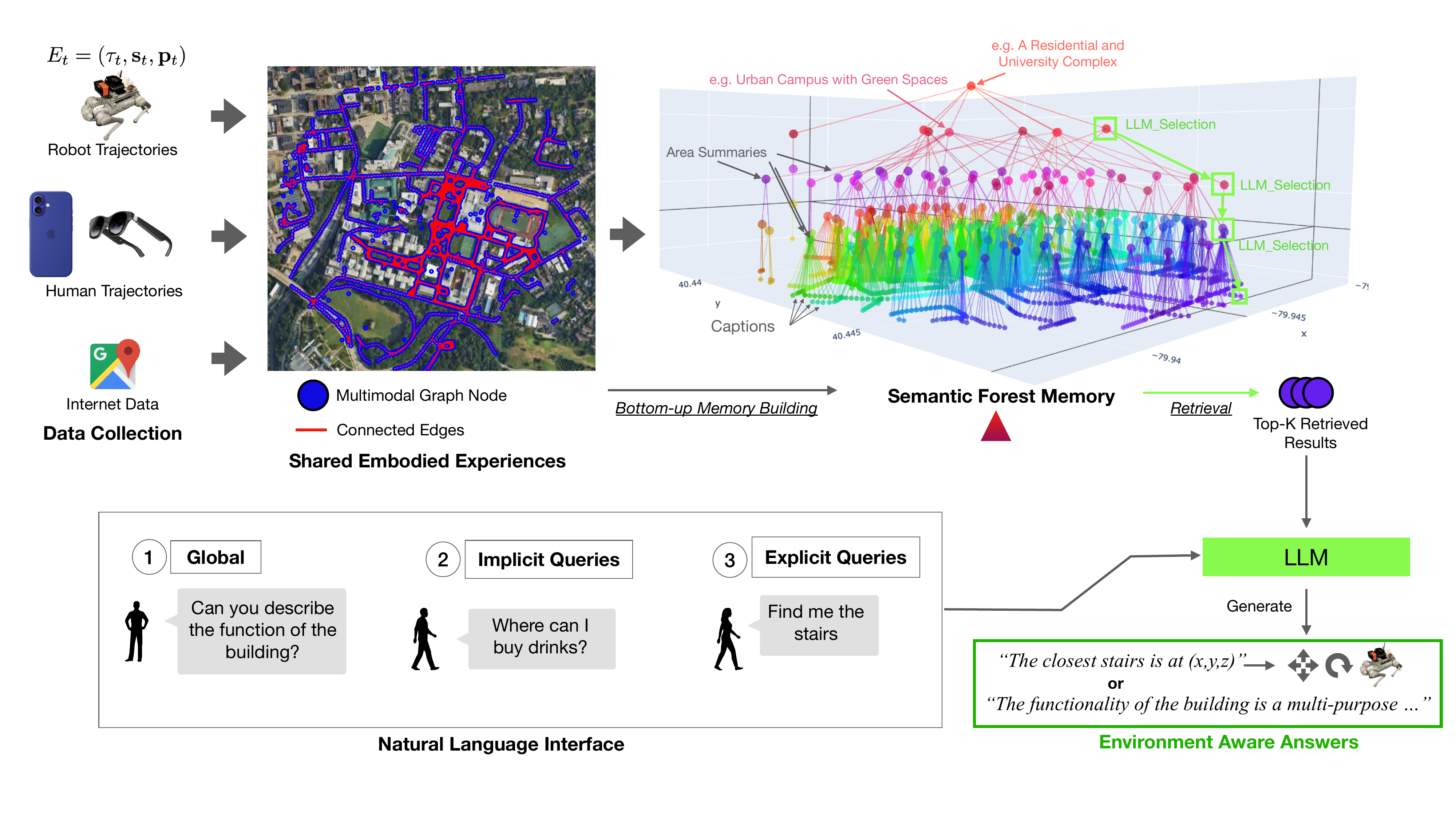} 
    \caption{\textbf{Overview}: Our goal is for robots to navigate and communicate effectively in any environment where humans are present. We introduce Embodied-RAG, a framework for automatically building hierarchical spatial memory and providing both explanations and navigation across multiple levels of query abstraction. Embodied-RAG supports robotic operations regardless of the query's abstraction level, the platform, or the environment.}
    
    \label{fig:figure1}
    \vspace{-1em}
\end{figure*}

\begin{abstract}
There is no limit to how much a robot might explore and learn, but all of that knowledge needs to be searchable and actionable. Within language research, retrieval augmented generation (RAG) has become the workhouse of large-scale non-parametric knowledge, however existing techniques do not directly transfer to the embodied domain, which is multimodal, data is highly correlated, and perception requires abstraction.

To address these challenges, we introduce Embodied-RAG, a framework that enhances the foundational model of an embodied agent with a non-parametric memory system capable of autonomously constructing hierarchical knowledge for both navigation and language generation. Embodied-RAG handles a full range of spatial and semantic resolutions across diverse environments and query types, whether for a specific object or a holistic description of ambiance. 
At its core, Embodied-RAG’s memory is structured as a semantic forest, storing language descriptions at varying levels of detail. This hierarchical organization allows the system to efficiently generate context-sensitive outputs across different robotic platforms. We demonstrate that Embodied-RAG effectively bridges RAG to the robotics domain, successfully handling over 250 explanation and navigation queries across kimometer-level environments, highlighting its promise for general-purpose non-parametric system for embodied agents.
\end{abstract}

% \keywords{Embodied Memory and Retrieval, Semantic Navigation, Retrieval Augmented Generation} 
\begin{IEEEkeywords}
Autonomous Agents, RAG, Embodied memory, Language-guided Robotics
\end{IEEEkeywords}

%===============================================================================

\section{Introduction}
\label{sec:intro}
% \input{text/1-intro}
\input{text/1-intro-new}

\section{Related Works}
\label{sec:relatedwork}

\input{text/3-relatedwork}

\vspace{-0.5em}

% % \section{Task: Embodied Generation}
% \label{sec:probleme}
% \input{text/2-problem-formulation}

\section{Method: Embodied Retrieval and Generation}
\label{sec:method}
\input{text/4-method}

\section{Experiments}
\label{sec:exp}
\input{text/5-exp}

%\subsection{Discussion}
\label{sec:discussions}

\input{text/6-discussions}
\section{Limitations and Future work}

\label{sec:Limiataions}
Embodied-RAG is not a drop-in replacement for other mapping approaches within the robotics community, but rather a supplement that focuses on open-world hierarchical semantics, building a high-level nonparametric memory system for easy integration of language models and related technologies with embodied agents versus low-level mapping and planning or precise vision reasoning often necessary in robotics.

For example, our work assumes access to a perfect local planner for the navigation task. This results in our system not guaranteeing robustness in obstacle avoidance involving dynamic objects and people. A natural question for future work is to include dynamic objects in the memory but this requires also reasoning over the concept of stale observations. 

On the topic of visual reasoning, Embodied-RAG struggles with requests that require precise counting of objects at a small scale (e.g., ``How many chairs are there around the red table?''). This limitation arises because the agglomerative clustering of the semantic forest does not consider multi-view consistency. Generally, current (V)LMs struggle with 3D spatial reasoning, so future work could need to explicitly incorporate multi-view consistency techniques into the hierarchies of the semantic forest with a learned or pre-trained mechanism to cluster with positional information (e.g. utilizing a LLM). 

Finally, the reliance on (V)LM APIs creates a deployment dependency of nearly uninterrupted internet access. While we tackled some of this directly in our efficiency evaluation, the question of what knowledge and abilities are lost when models are distilled and quantized to the point of being deployable offline on local compute is an open research topic with the NLP community and is left to future work. 
% Future work could change cloud based LLM to on device LLM to make Embodied-RAG applicable to internet denined areas.  

% Although we name our memory ``General'' Embodied Memory, its focus and contribution is on the semantic forest than the topological map. Thus, it may be not be robust to obstacle avoidance of dynamic objects and people, etc. 
% Furthermore, currently GEM-RAG is inaccurate for requests for precise counting objects at a small scale (how many chairs are there around the red table?) This is due to the fact that the agglomerative clustering of the semantic forest does not consider multiview consistency. Future work can consider how to consider multiview consistency with positional information, with methods such as including them in the LLM prompt.  

\section{Conclusions} 

\label{sec:conclusions}
\input{text/conclusion}
%===============================================================================

% Distinguish the multiview consistency, by giving positional information in the prompt. 

% Furthermore, although GEM-RAG has potentials for handling requests at a lifelong scale (before mapping, while mapping, after mapping), there currently does not exist a mechanism to update or edit the contents nodes of the semantic forest, when the environment changes across time. However, in future work, it is an intuitive approach to include this mechanism by asking method LLM's to update. 

% \section{Citations}
% \label{sec:citations}

% 	Citations can be made using either \textbackslash citep\{\} or \textbackslash citet\{\}, depending from the appropriateness. To avoid the citation moving to the next line, it is often a good practice to replace the space before with a tilde (\~{}) character.
% 	Example 1: ``CoRL is the best conference ever~\citep{Gauss1857}.''
% 	Example 2: ``\citet{Lagrange1788} proved, both theoretically and numerically, that CoRL is the best conference ever.''

%===============================================================================

\clearpage
% The acknowledgments are automatically included only in the final and preprint versions of the paper.
% \section*{Acknowledgment}
% This material is based upon work supported by the Defense Advanced Research Projects Agency (DARPA) under Agreement No. HR00112490375 and partially supported by funding from Lockheed Martin Corporation.

%===============================================================================

\bibliographystyle{IEEEtran} % or another appropriate style
\bibliography{example}

\clearpage

\end{document}

%% file: text/1-intro-new.tex
% The mechanisms of \textit{memory}, \textit{retrieval}, and \textit{generation} enable humans to function as generalist embodied agents. We log our perceived and acted experiences at various levels of resolution. From this memory, we can retrieve information in multiple formats, such as moments, locations, facts, and holistic feelings. To generate responses in language or action, we connect the retrieved result to match the abstraction of the request. This coupling of memory construction, retrieval, and generation allows us to handle diverse requests (e.g., “Can you guide me to the restroom?” or “Summarize the trend of vegetation on the farm”) at any \textit{resolution}, from specific to general, and across any \textit{output format}, whether action or language, in any environment where we operate.
\renewcommand\thefootnote{}\footnote{\href{https://quanting-xie.github.io/Embodied-RAG-web/}{https://quanting-xie.github.io/Embodied-RAG-web/}}

%\vspace{-1em}
% Humans's paramatric brain are not good at remember every details that happened in everyday life at a granular details. We cannot retrieve quickly or previsely to a specific location over a large environment or without using external memory or search tools. Environment understanding, 

It is difficult for the human mind to determine what information should be remembered from our perceptually rich lived experiences. Where details are necessary, we revisit an experience or build explicit external representations like maps to capture the intricacies. This process begs questions of what the right semantic level and context should be indexed. Robots are now in the same but opposite position.  While dense SLAM and metric maps can be constructed, they become intractable to scale, and they do not track with larger semantic categories we find most useful in human memory -- we discard almost all low-level information as redundant and easy to rediscover. 
%Humans' parametric brains are not good at remembering every detail that happens in everyday life at a granular level. We cannot quickly or precisely retrieve specific information from a large environment without using external memory or search tools. Embodied agents like robots can help us retrieve information if they can log embodied experiences and interact with us. However, current memory and retrieval systems for embodied experiences have some issues. For robots that build metric maps and project semantic information onto these maps, scalability is poor. The cost to build and retrieve from these maps is expensive, and due to the nature of embodied experiences, much of the information is redundant.

Within Natural Language Processing (NLP), Retrieval-Augmented Generation (RAG)\cite{rag1,rag2,rag3} integrates non-parametric memory into Large Language Models (LLMs), enabling the use of large text corpora as private knowledge bases to enhance a model's memory, relevance, and factual grounding of model outputs, particularly in scenarios requiring access to up-to-date or domain-specific knowledge. We ask if such insights can be leveraged to endow robots with better scaling semantic memory, and what new technologies need to be invented to handle embodied experiences.
%The up-to-date and domain-specific properties make this approach appealing for robotics tasks.

Applying RAG to robotics presents unique challenges due to key differences between textual data and embodied experiences. First, embodied experiences are multimodal -- How do we make such data queryable for a RAG system? Unlike Internet documents, which are distinct and well-structured text, embodied data often consist of tuples of time, sensor observations, and robot poses, $E_t = (\tau_t, \mathbf{s}_t, \mathbf{p}_t)$. These multidimensional data need to be efficiently coupled and stored. Further, current representations of embodied experiences, such as dense point-cloud maps, fail to abstract the relevant semantics needed for a natural language query. %capture the multimodal nature and are hard to interpret. 
Although 3D scene graphs\cite{Hughes2022HydraAR} are interpretable, they rely on human-engineered schemas %to identify rooms and cannot 
that do no scale to diverse outdoor environments.

\newcommand{\GC}{\textbf{\color{green}\checkmark}}

\begin{table}[t!]
    \centering
    \begin{tabular}{@{}lcccc@{}}
        %\hline
        \textbf{Category} & 
            \textbf{Metrics Maps} & 
            \multicolumn{1}{c}{\shortstack{\textbf{Semantic} \\ \textbf{Metric Maps}}} & 
            \textbf{RAG} & 
            \multicolumn{1}{c@{}}{\shortstack{\textbf{Embodied-} \\ \textbf{RAG}}} \\
        \toprule
        \textbf{\textit{Retrieval}} & & & & \\
        $P(A | Q)$       & \texttimes & \GC        & \GC        & \GC \\
        $P(A | L)$       & \GC        & \GC        & \texttimes & \GC \\
        $P(A | L, Q)$    & \texttimes & \GC        & \texttimes & \GC \\
        $P(A | S, Q)$    & \texttimes & \texttimes & \GC        & \GC \\
        $P(A | S, L, Q)$ & \texttimes & \texttimes & \texttimes & \GC \\
        \midrule
        \textbf{\textit{Generation}} & & & & \\
        Text & \texttimes & \texttimes & \GC & \GC \\
        Waypoint & \GC & \GC & \texttimes & \GC \\
        Path & \GC & \GC & \texttimes & \GC \\
        \bottomrule
    \end{tabular}
    \caption{Comprehensive comparison of Metrics Maps, Semantic Metric Maps, RAG, and Embodied-RAG frameworks in terms of retrieval and generation capabilities. Here, $Q$ represents the query, $L$ denotes the embodied agent's position, and $S$ refers to other sensor data.}
    \label{tab:comparison}
\end{table}
Second, Naive RAG\cite{rag1} lacks the cross-document structural awareness needed for building spatially informed % cannot be globally aware the environment since no structure is imposed on the data to create organized 
knowledge graphs, and structured graphical RAG methods\cite{graphrag,lightrag} are too inefficient to build and query for %and take too long to build, preventing 
real-time deployment. % of RAG memory.
Finally, embodied observations are redundant and repetitive, which can confuse the retriever when attempting to select the correct context using semantic similarity alone, requiring extra reasoning steps during inference. 

\hypersetup{
colorlinks=true, % enables colored links again
linkcolor=red,   % red clickable links
urlcolor=red,    % red URLs
pdfborder={0 0 1} % adds a border around clickable links
}

To address these challenges, we present Embodied-RAG. Embodied-RAG has two components: \textit{Bottom-up Memory Building} (Fig.\ref{fig:figure2}(a)) and \textit{Top-down Retrieval} (Fig.\ref{fig:figure2}(b, c)). During \textit{Bottom-up Memory Building}, we address the first two problems: multimodal representation and efficient integration of structure into embodied experiences. The system first represents embodied experiences with a multimodal topological graph, where each node contains robot poses, robot observations (images), and timestamps. Based on these topological nodes, a \textit{semantic forest} is hierarchically clustered based on spatial proximity. This graph-building process is \textit{7.38X} faster than Graph-RAG and \textit{9.76X} faster than Light-RAG on the same dataset size, and can be extended in real-time. This two-stage memory system creates an efficient, large-scale, globally aware, interpretable, and multimodal memory representation for embodied agents to retrieve from.

In the \textit{Top-down Retrieval} process, to overcome the third challenge, we enhance retrieval performance across three different query types (explicit, implicit, and global), outperforming the state-of-the-art RAG baselines\cite{rag1,lightrag, graphrag}. Instead of relying solely on semantic similarity, Embodied-RAG incorporates a robust reasoning component for retrieval. This involves parallelized tree traversals with a selection-LLM. This retrieval process uses abstracted information from the top nodes to guide retrieval to the correct lower nodes with equal probabilities. For example, a toothbrush node is more likely inside a bathroom node, and a bench in a backyard is less desirable than a bench in a park for quietly reading a book.

To evaluate Embodied-RAG, we present a new dataset called the \textbf{Embodied-Experiences Dataset} for Embodied-RAG tasks. It contains topological graphs collected from 14 photorealistic simulated and 5 real environments of varying scales paired with two hundred queries with ground truth labeled information.

Furthermore, our experiments demonstrate that Embodied-RAG is a more efficient graphical non-parametric memory for embodied data, surpassing the baselines\cite{lightrag,graphrag} in memory building time, and achieve better performance across all query types.  In addition, semantic forest is a more versatile memory, as shown in Table \ref{tab:comparison}, and it is capable of taking multi types of input, while also able to be applied to various forms of embodiment (drones, locobots, quadrupeds) as global planner, seamlessly integrated with existing low-level autonomous navigation pipelines. This highlights Embodied-RAG's potential as a general system capable of task-, environment-, and platform-agnostic operation, enabling robots to effectively navigate and communicate in any environment where humans are present.

\noindent The key contributions and implications of this paper include:
\begin{itemize}[noitemsep,leftmargin=15pt]
\item \textbf{Task:} We extend RAG into embodied settings and highlight the unique challenges of retrieval from embodied experiences.
\item \textbf{Dataset:} We present the \textit{Embodied-Experience Dataset}, formulating semantic navigation and question answering under a single paradigm (Table \ref{tab:comparison}, Figure \ref{fig:figure1}).
\item \textbf{Method:} We show an initial step toward solving the challenges in representing and retrieving from embodied experiences, outperforming Naive-RAG, GraphRAG\cite{graphrag}, and LightRAG\cite{lightrag} on different query types across 19 diverse real and simulated environments. In addition, the high-speed memory building process make it applicable in real-time navigation and mapping. 
\item \textbf{Implications:} Our results and discussion provide a basis for rethinking approaches to generalist robot agents based on language-form non-parametric memories.
\end{itemize}

%% file: text/3-relatedwork.tex
\subsection{Retrieval and Generation}
 Retrieval-Augmented Generation (RAG) systems integrate large language models (LLMs) with external text corpora to enhance factual grounding and relevance in generated outputs ~\cite{rag1, ram2023, fan2024surveyragmeetingllms, gao2024rag}. Traditional RAG models embed user queries and document chunks into a shared vector space, retrieving the top-k most semantically similar text fragments to augment the model's context window ~\cite{gao2022precisezeroshotdenseretrieval, chan2024rqraglearningrefinequeries }. This approach effectively improves performance on tasks requiring domain-specific or up-to-date information. However, naive RAG systems\cite{gao2024rag} rely heavily on fragmented text chunks and simple similarity-based retrieval, limiting their ability to capture comprehensive and globally coherent information. Advanced RAG models such as GraphRAG~\cite{graphrag} and LightRAG\cite{lightrag} have been developed to overcome these limitations by extract entities and their relationships, organize them into graph structure for more complete and globally aware retrieval. However, due to the intrinsic nature of embodied experiences are often redundant, hierarchically correlated, and spatially grounded, these purely textal graph building approaches don't perform that well. In contrast, Embodied-RAG utilized spatial correlations to build spatially related scene graphs.

\subsection{Existing Methods of Semantic Memory and Retrieval}
Several methods have been proposed for storing and querying semantic memory in spatial environments, but they remain limited and task-specific compared to the potential of foundation models. Approaches like \cite{min2021film, min2022don, chaplot2020object} associate voxels with predefined object categories, enabling fixed vocabulary retrieval, while methods such as \cite{shafiullah2022clipfields, huang23vlmaps} map voxels to image embeddings, allowing for open vocabulary queries. Systems like \cite{chang2023goat} store images per voxel, supporting queries about people, language/image inputs, and object categories. However, a common challenge across these approaches is aligning the semantic abstraction with the spatial resolution. Queries such as ``cup,'' ``red cup,'' or ``I want to heat my lunch'' are object-level, but methods like \cite{ramakrishnan2022poni, min2023self} focus primarily on local retrieval during exploration, using structured frontiers based on object layouts. Scene graphs \cite{scenegraph, sayplan}, while free from dense memory issues, rely on human-engineered schemas (e.g. floor $\xrightarrow{}$ room $\xrightarrow{}$ object $\xrightarrow{}$ asset), making them unsuitable for novel or outdoor environments.

Other approaches, such as OCTREE maps \cite{hornung2013octomap} and their semantic versions \cite{zhang2018semantic, zheng2023asystem, liu2019object}, organize occupancy data efficiently but still limit semantics to the object level. Methods like Semantic OCTREE \cite{zhang2018semantic, liu2019object} and GENMos \cite{zheng2023asystem} use fixed object categories, lacking support for free-form language queries or varying levels of spatial and semantic resolution needed for holistic understanding.

\subsection{Semantic Navigation and Question Answering}
Tasks like ObjectNav \cite{chaplot2020object, min2023self, anderson2018evaluation}, ImageNav \cite{imagenav, zhu2017target, krantz2022instance}, and Visual Language Navigation \cite{vlnsurvey} assess a robot’s ability to navigate towards semantic targets based on object categories, images, or language descriptions. While recent efforts like GOATBench \cite{khanna2024goat} combine multiple input types, these tasks still focus on object-level queries and lack the flexibility to handle broader, more abstract user requests. Embodied Question Answering (EQA) \cite{das2018embodied, padalkar2023open, yu2019multi, tan2023knowledge} and Video Question Answering (VideoQA) \cite{zhong2022video, yang2003videoqa, castro2020lifeqa, xiao2021next} extend navigation by requiring text-based answers within actionable or video environments, though EQA is limited to indoor settings and VideoQA lacks active navigation. Our approach expands these paradigms by integrating action-based and question-answering capabilities across a wider range of environments and user queries.

%% file: text/4-method.tex
\begin{figure*}[!t]
    \centering
    \includegraphics[scale=0.27,trim={0 15cm 0 0},clip]{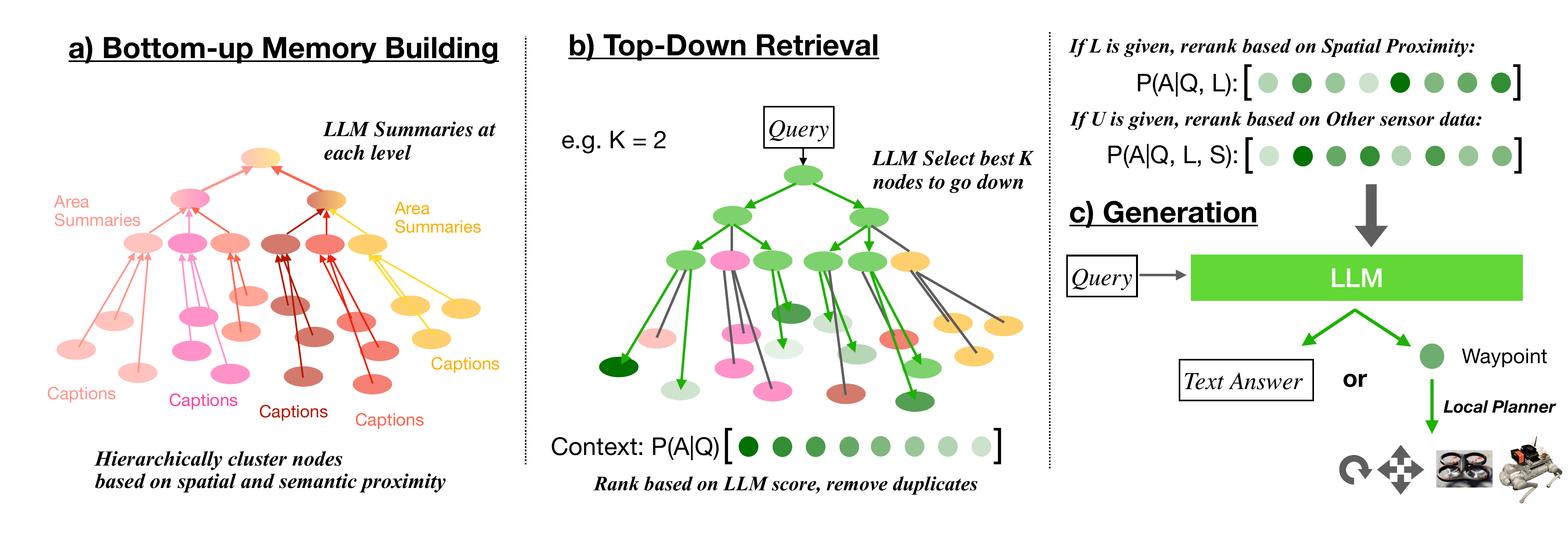}
    %\vspace{-10em}
    \caption{\textbf{Embodied-RAG method overview}. (a) Memory is constructed by hierarchically organizing the nodes of the topological map into a \textit{semantic forest}. (b) The memory in (a) can be retrieved for a query, with parallelized tree traversals. (c) Navigation actions with text outputs, or global explanations can be generated for the query, with using the retrieval results as LLM contexts.}
    \label{fig:figure2}
    %\vspace{-1em}
\end{figure*}

\subsection{Bottom-up Memory Construction}
\label{sec:graph_frontier}
% \subsubsection{Memory-Action Representation}
\noindent The memory construction process of Embodied-RAG consists of two parts: a topological map and a \textit{semantic forest}. \\[-10pt]

\noindent\textbf{Topological Map}
We employ a topological graph composed of nodes with the following attributes:
\begin{itemize}[noitemsep,leftmargin=15pt]
% \vspace{-5pt}
    \item Pose information: The (x, y, z) position and yaw angle $\theta$ on the map where the image was captured. Blue nodes in \ref{fig:figure1} are the topological nodes, and they are connected according to agent's path history or within a threshold $\alpha$ 
    \vspace{5pt}
    \item Timestamps
    \vspace{3pt}
    \item Images: Ego-centric images.
    \vspace{3pt}
    \item Captions: Generated by a VLM (GPT-4o), these captions provide detailed textual descriptions of the image.  
    % \vspace{-5pt}
\end{itemize}

The nodes form a topological map (blue nodes in Fig. \ref{fig:figure2}), eliminating the need for specific control parameters like velocity and yaw, which often vary across different drive systems. This abstraction enables compatibility with any local planner, regardless of the robot's embodiment. Furthermore, the topological structure is far more memory-efficient than traditional metric maps \cite{chaplot2020object, shafiullah2022clipfields, min2021film}, allowing for efficient scaling in both large outdoor and complex indoor environments. Our experiments show that this approach successfully operates on kilometer-scale topological graphs. 

\noindent\textbf{Semantic Forest}  The concept of a semantic forest leverages the observation of intrinsic structure of embodied data, where objects and scenes naturally exhibit spatial and semantic organization. By capturing higher-level spatial and semantic information in a hierarchical tree structure, known as a \textit{semantic forest}, we can effectively model these relationships through a two-step iterative process: clustering and summarization.

First, we employ complete-linkage hierarchical clustering (CLINK) \cite{sneath1973numerical, Mllner2011ModernHA} with a novel hybrid distance metric to group leaf nodes (level 0 nodes in Fig.~\ref{fig:figure2}(a)). The hybrid similarity matrix computation integrates both spatial and semantic relationships between nodes through a weighted combination approach. The similarity ($S_{ij}$) of two nodes \(i\) and \(j\) is defined as the weighting of two terms: % computed as:

\[
S_{ij} = (1-\alpha) S^{\text{spatial}}_{ij} + \alpha S^{\text{semantic}}_{ij}
\]

The components of the similarity metric are as follows:

\subsubsection*{Spatial Similarity \((S_{ij}^{\text{spatial}})\)}
The spatial similarity is computed using the haversine distance with exponential decay:

\[
S_{ij}^{\text{spatial}} = \exp\left(-\frac{d_{\text{haversine}}(i, j)}{\theta}\right)
\]

Where \(d_{\text{haversine}}(i, j)\) is the great-circle distance between locations \(i\) and \(j\), and \(\theta\) is the base distance threshold.
%\begin{itemize}
%    \item \(d_{\text{haversine}}(i, j)\): The great-circle distance between locations \(i\) and \(j\).
%    \item \(\theta\): The base distance threshold.
%\end{itemize}

\subsubsection*{Semantic Similarity \((S_{ij}^{\text{semantic}})\)}
The semantic similarity is computed as the cosine similarity between neural language embeddings \(\mathbf{e}_i, \mathbf{e}_j\) of the text descriptions for nodes \(i\) and \(j\).

\[
S_{ij}^{\text{semantic}} = \frac{\mathbf{e}_i \cdot \mathbf{e}_j}{\|\mathbf{e}_i\| \|\mathbf{e}_j\|}
\]

%\begin{itemize}
    %\item \(\mathbf{e}_i, \mathbf{e}_j\): Neural embeddings of the textual descriptions for nodes \(i\) and \(j\).
%\end{itemize}

Once the clusters are formed at each level, we generate semantic summaries for each cluster using an LLM summarizer (e.g., GPT-4). The summary and the average distance between nodes are saved as new nodes (level 1-3 nodes in Fig.~\ref{fig:figure2}(a)). This bottom-up clustering process continues until either root nodes are formed or no further meaningful clusters can be created. The summarization process is parallelized across clusters at the same hierarchical level, ensuring efficient processing of the entire forest structure. 

Unlike traditional 3D Scene Graph approaches~\cite{scenegraph, Hughes2022HydraAR}, which often rely on manual rules for identifying rooms and functional spaces, our hierarchical structure and corresponding semantic trees enable the automatic creation of meaningful semantic regions. This approach is particularly advantageous for outdoor navigation, where walls and physical structures are absent and cannot be used to infer functional areas.

\subsection{Top-down Retrieval}
\label{sec:retreival}
% \begin{figure*}[ht]
%     \centering
%     \includegraphics[scale=0.37]{images/Figure3.png}
%     \vspace{-0.25em}
%     \caption{We illustrate three retrieval methods: (a) Semantic Match, (b) RAG, and (c) our proposed method, Embodied-RAG. Semantic Match retrieves the node in the topological map with the highest cosine similarity with the query, while RAG retrieves top $k$ most semantic similar nodes. In contrast, Embodied-RAG retrieves the LLM reasoning paths. The gray icons represent different levels of information abstraction: the black square at the bottom indicates the most basic and unprocessed information, while the black squares higher up represent progressively more abstracted and generated information. Embodied-RAG leverages these abstractions to semi-attend to base nodes that are not explicitly retrieved, enhancing the generation model's understanding of the global structure and relationships among nodes.}
%     \label{fig:figure4_retrieval}
%     \vspace{-1em}
% \end{figure*}
% \vspace{-1em}

To address the challenges posed by redundant and repetitive embodied observations, which make retrieval difficult when relying solely on semantic similarities as in Naive RAG \cite{rag1}, and to enhance reasoning capabilities over hierarchies of abstraction constructed for a given environment, we modified RAG's relevancy scoring mechanism  in a manner inspired by Tree-of-Thoughts \cite{yao2024tree}. Specifically, the two phase retrieval mechanism transitions from semantic similarity to LLM-based selections at each hierarchical level:
\subsubsection*{Phase 1: Semantic-Guided Hierarchical Traversal}

The first phase involves a top-down parallel exploration of the semantic forest, where node selection is guided purely by semantic relevance. For a given query \(q\), we define a selection function:

\[N_l = f_{\text{LLM}}(q, C_l, k)\]

Here, \(N_l\) represents the set of selected nodes at level \(l\), \(C_l\) denotes the candidate nodes at that level, and \(f_{\text{LLM}}\) is the LLM-based selection function. The parameter \(k\) specifies the branching factor. The algorithm recursively explores the children of each selected non-base node until the base nodes are reached. By focusing on semantic relevance during traversal, this phase prunes irrelevant branches early and ensures computational efficiency, while still comprehensive exploration over the entire semantic forest.

\subsubsection*{Phase 2: Hybrid Re-ranking}

Once the base nodes are collected, they are scored individually by another LLM and ranked in descending order of relevance, kept in a set to remove duplicates. If location information is provided, the probability distribution is updated from \(P(A \mid Q)\) to \(P(A \mid Q, L)\). A re-ranking based on spatial proximity is then conducted by computing spatial scores and combining them with semantic scores, weighted by a factor \(\alpha\). The spatial scores are calculated in the same manner as the spatial similarity described in Section~\ref{sec:graph_frontier}.

Additionally, if the embodied experiences include more sensor data, these data can be preprocessed to refine the scoring mechanism. For instance, in this study, Normalized Difference Vegetation Index (NDVI) data is used to determine the quality of the grass. This score is integrated into the re-ranking process to prioritize regions with low-quality grass. The preprocessing and re-ranking steps can be adapted to the specific requirements and its operational goals, allowing the framework to remain flexible across different applications.

\subsection{Generation}

The retrieved nodes are passed as part of the context, along with the user query, to a generation LLM. The LLM is prompted to handle two types of queries: (1) For \textit{``find"} queries (explicit or implicit), it outputs a desired waypoint in a JSON format, along with reasoning for its choice; (2) For \textit{``explain"} queries, it generates a textual response. The detailed prompt used for the generation process is available on our project website.
The Embodied-RAG pipeline can be conceptualized as a global planner for \textit{``find"} queries. Since the topological graph contains connectivity information, a Dijkstra’s path planning algorithm is employed to compute the minimum-distance path between the current location and the selected waypoint. For navigation between waypoints, any local planner can be integrated. In this paper, we use the Unitree-Go2's ``go-to-waypoint" API as our local planner.

%% file: text/5-exp.tex
%\subsection{Setup}

\subsection{Embodied-Experiences Dataset}

The datasets utilized in the Embodied-RAG task are structured as topological graphs (see Section~\ref{sec:method} for node details).

The dataset is composed of diverse environmental settings collected through complementary data collection techniques. Real-world environments were explored using autonomous robots to construct three detailed indoor graphs and one mixed outdoor-indoor graph, capturing realistic navigation scenarios. To model large-scale urban spaces, a comprehensive street-view graph was created using imagery from Google Street View, providing broad and complex spatial data. Additionally, fourteen object-centric topological graphs were generated using the photo-realistic AirSim~\cite{shah2018airsim} simulator, enabling controlled simulation of varied and intricate environments. On average, the topological graphs contain approximately 50 nodes, reflecting moderate complexity in most environments, while the large-scale street-view graph consists of 3,525 nodes, offering extensive spatial coverage for evaluating more complex navigation and retrieval tasks.

The dataset is further divided based on modality for experimentation. In the \textit{E-image} setting, each topological graph, denoted as \(E_t = (\tau_t, \mathbf{s}_t, \mathbf{p}_t)\), includes nodes where \(\mathbf{s}_t\) contains only image data. In contrast, the \textit{E-multimodal} setting incorporates nodes where \(\mathbf{s}_t\) includes both images and additional sensory data. In the experiments shown in Table~\ref{table:performance_multimodal}, \(\mathbf{s}_t\) contains both image data and NDVI readings, reflecting the multimodal nature of the environment.

\subsection{Embodied-RAG Task}

\noindent We include %An Embodied-RAG task consists of 
two query types: \textit{Find} and \textit{Explain}. 

\textit{Find} queries has two subcategories: (1) Explicit Queries. These involve searching for a specific object instance or a clearly defined target (e.g., ``Find a bench"), (2) Implicit Queries: These require a more nuanced, pragmatic understanding, such as assessing adequacy or interpreting instructions with contextual reasoning (e.g., ``Find a quiet spot to read").  

For \textit{Explain} queries, the request may pertain to global information, such as describing a specific location or providing a general understanding of the environment (e.g., ``What's the vegetation trend of this environment?").

Example tasks are shown in Fig. \ref{fig:figure1}. \textit{Find} queries queries are navigational tasks that expect navigation actions and text descriptions of the retrieved location. \textit{Explain} queries are QA tasks requiring text generation at a more holistic level.

The queries were collected by four human annotators familiar with the Embodied-Experience datasets. The annotators created queries by reviewing the dataset's images and leveraging their understanding of the environmental context.

\section{Results}

\subsection{Evaluation}

To comprehensively assess the system's performance, we employ distinct evaluation metrics tailored to the nature of the queries: \textit{Find} and \textit{Explain}.

\subsubsection{\textit{Find Queries}}
To effectively evaluate the system's capabilities as a global planner, we separate \textit{navigation success} from \textit{generation success}. The system outputs an image path as the result and calculates the probability \( P(Q \mid A) \), representing the likelihood of finding the queried object given the generated answer (image path). This probability is determined using a \textbf{cross-voting technique} among five Vision-Language Models (VLMs), ensuring unbiased scoring for open-ended queries. 

Instead of binary checks, we use probabilities to account for the inherent ambiguity in implicit queries (e.g., ``Find me a place to eat''), where deterministic answers are not always feasible. Additional details about the VLM prompts and evaluation methodology are available on our project website.

If \textit{location information} is provided \( P(Q \mid A, L) \), we extend the evaluation by weighting \( P(Q \mid A) \) by the \textit{path length}, similar to established metrics like Success Weighted by (normalized inverse) Path Length (SPL)~\cite{anderson2018evaluation}. Specifically:
\[
P(Q \mid A, L) = P(Q \mid A) \times \frac{\text{path length}}{\text{radius of the environment}}
\]
This adjustment ensures the evaluation reflects not only the success of finding the object but also the efficiency of the navigation path.

\subsubsection{\textit{Explain Queries}}
For explanation queries, we constructed a \textbf{golden dataset} by collecting answers from expert annotators for each query. The system's generated responses are evaluated by computing the \textit{semantic similarity} between the generated answers and the corresponding golden answers \({SS(A, A_e)} \) A\_e represents the expert-provided answer.

\subsection{Baselines}

To evaluate our Embodied-RAG approach, we conducted comparative experiments against three baseline methods: \textit{Naive-RAG}, \textit{GraphRAG}, and \textit{LightRAG}. %Below, we describe each baseline in detail:

%The first baseline is \textit{Naive-RAG} \cite{rag1}. 
For compatibility with Naive-RAG \cite{rag1}, we converted the graph files from the Embodied-Experience dataset into plain text (\texttt{.txt}) files. The dataset is divided into text chunks, and the system retrieves semantically relevant chunks to populate the context window of GPT-4o (with a token limit of 16k). These retrieved chunks are used to generate enhanced responses. Naive-RAG does not leverage any structural knowledge of the dataset, treating it as flat text.

\textit{GraphRAG} \cite{graphrag} incorporates graph structures into the Naive-RAG system. After preprocessing the dataset into text chunks (similar to Naive-RAG), GraphRAG utilizes an LLM to extract entities and relationships from the text and aggregates them into different communities. A graph is then constructed to capture global relationships, with community reports summarizing the entities and their connections. During retrieval, GraphRAG generates multiple intermediate answers in parallel, one for each chunk, ranks them based on a helpfulness score, and iteratively adds the most relevant answers to the context window until the token limit is reached. The final answer is generated based on this enriched context.

Finally, \textit{LightRAG} \cite{lightrag} is a state-of-the-art graphical RAG approach designed for efficiency. LightRAG's contribution is more efficient retrieval by % processes the dataset similarly but retrieve more efficient. It 
indexing the graph using a dual-level key system (low-level and high-level keys). During retrieval, this dual-level retrieval method is used to retrieve relationships between key entities. Like the other baselines, the same preprocessing steps were applied to the dataset.

\begin{table}[t]
\centering
\begin{tabular}{
>{\centering\arraybackslash}p{0.6cm} l 
>{\raggedright\arraybackslash}p{0.9cm} % Input types column aligned left
>{\centering\arraybackslash}p{0.5cm} 
>{\centering\arraybackslash}p{0.5cm} 
>{\centering\arraybackslash}p{0.5cm} 
>{\centering\arraybackslash}p{1 cm}}
\toprule
\textbf{Query Types} & \textbf{Metrics} & 
\textbf{Input Types} & 
\textbf{Naive-RAG} & 
\textbf{Graph-RAG} & 
\textbf{Light-RAG} & 
\textbf{Embodied-RAG} \\
\midrule
\multirow{2}{*}{\textbf{Explicit}} 
& \textbf{P(Q$\mid$A)}\(\uparrow\) & Q only & 0.08  & 0.06  & 0.08  & \textbf{0.55} \\
& \textbf{P(Q$\mid$A, L)}\(\uparrow\) & Q, L  & 0.041  & 0.029  & 0.027  & \textbf{0.28} \\
\midrule
\multirow{2}{*}{\textbf{Implicit}} 
& \textbf{P(Q$\mid$A)}\(\uparrow\) & Q only & 0.10  & 0.12  & 0.13  & \textbf{0.62} \\
& \textbf{P(Q$\mid$A, L)}\(\uparrow\) & Q, L  & 0.07  & 0.06  & 0.07  & \textbf{0.25} \\
\midrule
\multirow{1}{*}{\textbf{Global}} 
& \textbf{SS(A, A\_e)}\(\uparrow\) & Q only & 0.31  & \textbf{0.68} & 0.65  & 0.67 \\
\bottomrule
\end{tabular}
\caption{Performance Comparison on the \textit{E-image} Dataset across different methods. Embodied-RAG consistently outperforms all other methods across Explicit, Implicit, and Global query types, achieving the highest scores in both retrieval probabilities (\( P(Q \mid A) \) and \( P(Q \mid A, L) \)) and semantic similarity (\( SS(A, A_e) \)) metrics. Input types used in evaluation are specified for each query type.}
\label{table:performance-image}
\end{table}

\begin{table}[t]
\centering
\begin{tabular}{
>{\centering\arraybackslash}p{0.6cm} l 
>{\raggedright\arraybackslash}p{0.9cm} 
>{\centering\arraybackslash}p{0.5cm} 
>{\centering\arraybackslash}p{0.5cm} 
>{\centering\arraybackslash}p{0.5cm} 
>{\centering\arraybackslash}p{0.9 cm}}
\toprule
\textbf{Query Types} & \textbf{Metrics} & 
\textbf{Input Types} & 
\textbf{Naive-RAG} & 
\textbf{Graph-RAG} & 
\textbf{Light-RAG} & 
\textbf{Embodied-RAG} \\
\midrule
\multirow{3}{*}{\textbf{Explicit}} 
& \textbf{P(Q$\mid$A)}\(\uparrow\) & Q only & 0.08  & 0.09 & 0.12 & \textbf{0.58} \\
& \textbf{P(Q$\mid$A, L)}\(\uparrow\) & Q, L  & 0.03  & 0.04 & 0.03 & \textbf{0.28} \\
& \textbf{P(Q$\mid$A, L)}\(\uparrow\) & Q, L, S  & 0.04  & 0.04 & 0.03 & \textbf{0.36} \\
\midrule
\multirow{3}{*}{\textbf{Implicit}} 
& \textbf{P(Q$\mid$A)}\(\uparrow\) & Q only & 0.10  & 0.12 & 0.13 & \textbf{0.67} \\
& \textbf{P(Q$\mid$A, L)}\(\uparrow\) & Q, L  & 0.04  & 0.05 & 0.07 & \textbf{0.29} \\
& \textbf{P(Q$\mid$A, L)}\(\uparrow\) & Q, L, S & 0.04  & 0.04 & 0.08 & \textbf{0.41} \\
\midrule
\multirow{2}{*}{\textbf{Global}} 
& \textbf{SS(A, A\_e)}\(\uparrow\) & Q only & 0.60 & 0.72 & \textbf{0.75} & 0.74 \\
& \textbf{SS(A, A\_e)}\(\uparrow\) & Q, S & 0.46 & 0.68 & 0.78 & \textbf{0.95} \\
\bottomrule
\end{tabular}
\caption{Performance on the \textit{E-multimodal} Dataset with Input Types. This table includes sensor data as an additional input, represented in rows with \( Q, L, S \) or \( Q, S \). The results show that Embodied-RAG achieves significant performance improvements over metrics \( P(Q \mid A, L) \) when sensor data is incorporated. Embodied-RAG consistently outperforms other methods across all query types and metrics.}
\label{table:performance_multimodal}
\end{table}

\subsection{Quantitative Results}

The main results are presented in Table~\ref{table:performance-image} and Table~\ref{table:performance_multimodal}, where we evaluate performance on both the \textit{E-image} and \textit{E-multimodal} datasets using three different query types and three different input types. 
%
%As shown in both tables, 
The performance of the baselines is notably poor for explicit and implicit query types. This is primarily because chunking multimodal embodied data into text often fails to retrieve the correct images, resulting in retrieval failures in most cases. For global queries, however, the baselines \textit{LightRAG} and \textit{GraphRAG} outperform \textit{Naive-RAG}, demonstrating the effectiveness of graph structures in generating holistic responses about the environment.

Notably, Embodied-RAG outperforms all baselines for explicit and implicit queries across all input types. For global queries, particularly under \(P(A \mid Q, L)\), Embodied-RAG shows superior performance, highlighting its flexibility in the hybrid re-ranking step described in Section~\ref{sec:retreival}. This flexibility enables it to adapt actively to spatial constraints during retrieval.

In Table~\ref{table:performance_multimodal}, we observe that when sensor data is provided to Embodied-RAG, its performance further improves, while the baselines remain unaffected. This result emphasizes the advantages of integrating multimodal information, offering insights into how sensor data enhances the system's ability to understand and respond effectively to complex queries.

\begin{figure}[!tt]
    \centering
    \includegraphics[width=0.45\textwidth]{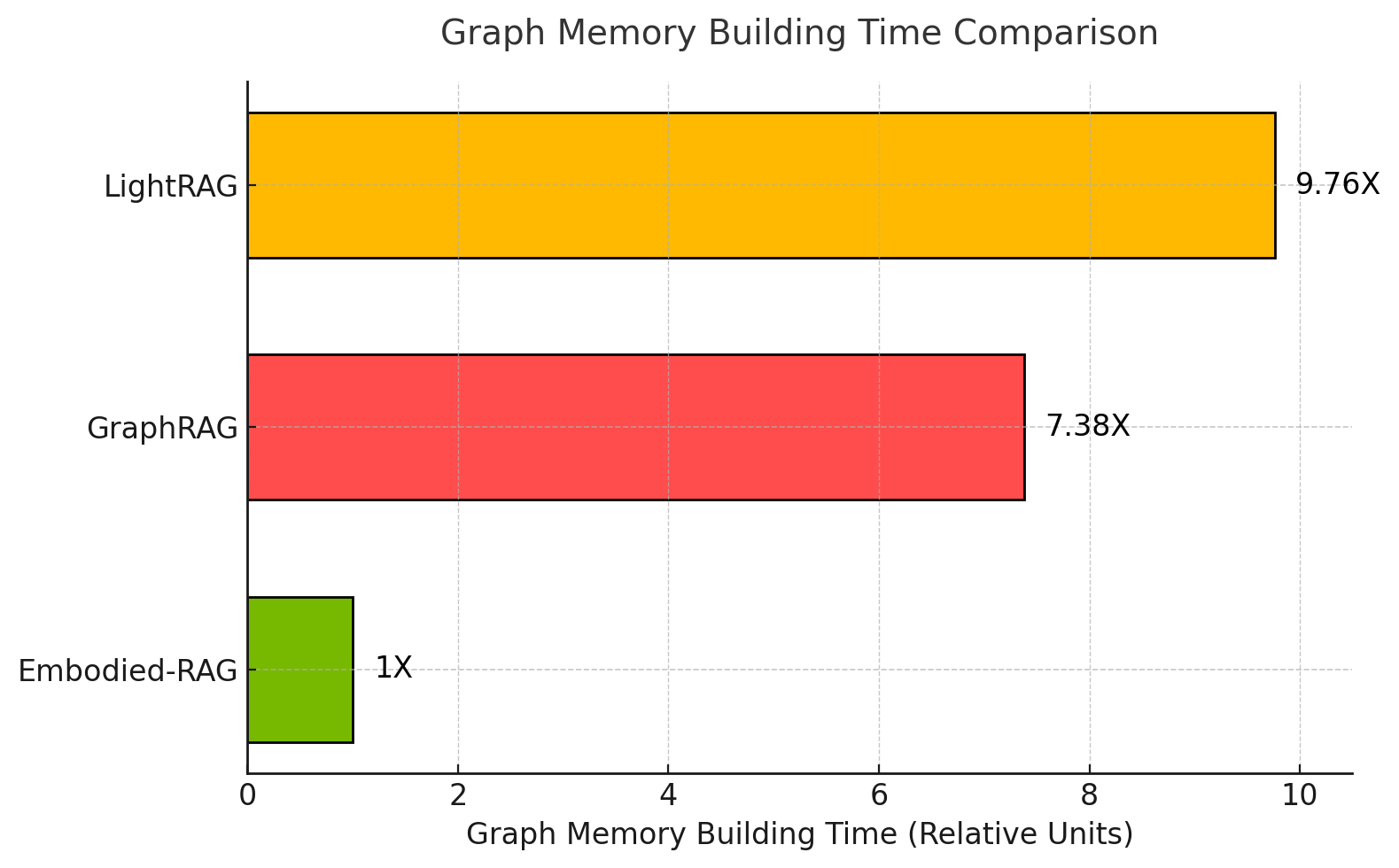}
    \caption{Graph Memory Building Time Comparison. The relative graph memory building time for Embodied-RAG, GraphRAG, and LightRAG is displayed, normalized to Embodied-RAG (1X). Embodied-RAG demonstrates significantly faster graph construction, being 7.38 times faster than GraphRAG and 9.76 times faster than LightRAG}
    \label{fig:time}
    %\vspace{-1em}
\end{figure}

% \begin{figure}[!tt]
%     \centering
%     \includegraphics[width=0.48\textwidth]{images/graph_building_time.png}
%     \caption{Memory retrieval time comparison}
%     \label{fig:time}
%     \vspace{-1em}
% \end{figure}

\subsection{Computation Results}

We conducted a computational comparison of the graphical memory building time between Embodied-RAG, \textit{LightRAG}, and \textit{GraphRAG}. The results, shown in Fig.~\ref{fig:time}, demonstrate that Embodied-RAG's graph-building process is 7.38 times faster than GraphRAG and 9.76 times faster than LightRAG. This efficiency is attributed to Embodied-RAG's semantic forest design, which leverages the inherent properties of embodied data. Specifically, objects or locations that are spatially close can naturally be abstracted into higher-level clusters, reducing the need for excessive LLM calls to generate relationships between individual text chunks as the baseline do. By clustering larger groups of information at once, Embodied-RAG creates a leaner graph structure.

On average, Embodied-RAG takes approximately 4 minutes and 35 seconds to build a complete semantic forest for a one-kilometer radius environment (e.g., the CMU dataset inside Embodied-Experiences dataset), consisting of 3,353 nodes. Furthermore, the semantic forest can be built incrementally by progressively clustering nodes at each hierarchical level.

\subsection{Ablation}

Since traditional RAG systems are not designed to handle embodied experience data, baselines perform poorly on \textit{Find} tasks. This raises the question as to whether the superior performance of Embodied-RAG is due to its memory structure or its retrieval mechanism. To explore this, we modified the baselines to retrieve directly from our \textit{semantic forest} memory, allowing a direct comparison of retrieval performance.
For each method, we computed \( P(Q \mid A) \) and path-weighted \( P(Q \mid A, L) \) for the top-1 and top-5 retrieved images. The results reveal that while our retrieval method achieves similar performance on \( P(Q \mid A) \), it consistently produces shorter path lengths. Consequently, Embodied-RAG demonstrates superior performance on \( P(Q \mid A, L) \), underscoring its ability to effectively leverage both semantic and spatial relationships within the semantic forest to build better context for generation.

\label{app:ablation}
\begin{table*}[t]
\centering
\begin{tabular}{l 
>{\centering\arraybackslash}p{1.5cm} 
>{\centering\arraybackslash}p{1.5cm} 
>{\centering\arraybackslash}p{1.5cm} 
>{\centering\arraybackslash}p{1.5cm} 
>{\centering\arraybackslash}p{1.5cm} 
>{\centering\arraybackslash}p{1.5cm} 
>{\centering\arraybackslash}p{1.5cm} 
>{\centering\arraybackslash}p{1.5cm}}
\toprule
\multicolumn{1}{c}{\textbf{Metrics}} & 
\multicolumn{2}{c}{\textbf{Naive-RAG}} & 
\multicolumn{2}{c}{\textbf{Graph-RAG}} & 
\multicolumn{2}{c}{\textbf{Light-RAG}} & 
\multicolumn{2}{c}{\textbf{Embodied-RAG}} \\
\cmidrule(lr){2-3} \cmidrule(lr){4-5} \cmidrule(lr){6-7} \cmidrule(lr){8-9}
 & Top-1 & Top-5 & Top-1 & Top-5 & Top-1 & Top-5 & Top-1 & Top-5 \\
\midrule
\multicolumn{9}{c}{\textbf{Explicit Queries}} \\
\midrule
\textbf{P(Q$\mid$A)}  &  0.51  & 0.55 & 0.64 & \textbf{0.522}  & 0.53 & 0.43 & \textbf{0.65} & 0.44 \\
\textbf{NI. Path Length}  & 0.38  & 0.47 & 0.44 & 0.45 & 0.39 & 0.37 & \textbf{0.48}& \textbf{0.68} \\
\textbf{P(Q$\mid$A, L)}  & 0.19  & 0.26 & 0.28 & 0.23 & 0.20 & 0.16 & \textbf{0.31} & \textbf{0.30} \\
\midrule
\multicolumn{9}{c}{\textbf{Implicit Queries}} \\
\midrule
\textbf{P(Q$\mid$A)}  &  0.60  & 0.51 & \textbf{0.66} & 0.60  & 0.64 & \textbf{0.60} & 0.62 & 0.57 \\
\textbf{NI. Path Length}  & 0.32  & 0.45 & 0.65 & 0.57 & 0.28 & 0.27 & \textbf{0.69} & \textbf{0.68} \\
\textbf{P(Q$\mid$A)}  & 0.19  & 0.23 & 0.38 & 0.34 & 0.18 & 0.16 & \textbf{0.41} & \textbf{0.39} \\
\bottomrule
\end{tabular}
\caption{Retrieval performance comparison for explicit and implicit queries on Embodied-RAG graph memory with \( P(A \mid Q, L) \) input. Metrics include \( P(Q \mid A) \), normalized inverse path length (NI. Path Length), and \( P(Q \mid A, L) \), evaluated for Top-1 and Top-5 results. Embodied-RAG consistently outperforms other methods across most metrics, particularly in \( P(Q \mid A, L) \) for both query types, showcasing its strong retrieval and navigation capabilities.}
\label{table:combined-top1-top5}
\end{table*}

% \begin{table*}[t]
% \centering
% \begin{tabular}{l 
% >{\centering\arraybackslash}p{1.5cm} 
% >{\centering\arraybackslash}p{1.5cm} 
% >{\centering\arraybackslash}p{1.5cm} 
% >{\centering\arraybackslash}p{1.5cm}  
% >{\centering\arraybackslash}p{1.5cm} 
% >{\centering\arraybackslash}p{1.5cm} 
% >{\centering\arraybackslash}p{1.5cm} 
% >{\centering\arraybackslash}p{1.5cm} }
% \toprule
% \multicolumn{1}{c}{\multirow{2}{*}{\textbf{Metrics}}} & 
% \multicolumn{4}{c}{\textbf{Explicit}} & 
% \multicolumn{4}{c}{\textbf{Implicit}} \\
% \cmidrule(lr){2-5} \cmidrule(lr){6-9}
% & Naive-RAG & Graph-RAG & Light-RAG & Embodied-RAG 
% & Naive-RAG & Graph-RAG & Light-RAG & Embodied-RAG \\
% \midrule
% \textbf{P(Q$\mid$A)}  & 0.931 & 0.896 & 0.931 & \textbf{0.965} & 0.804  & 0.882 & 0.86 & \textbf{0.96} \\
% \textbf{NI. Path Length}  & 0.44  & 0.48 & 0.52 & \textbf{0.54} & 0.38  & \textbf{0.68} & 0.58 & \textbf{0.68} \\
% \textbf{P(Q$\mid$A, L)}  & 0.44  & 0.48 & 0.52 & \textbf{0.54} & 0.38  & \textbf{0.68} & 0.58 & \textbf{0.68} \\

% \midrule
% % \textbf{Total}  & - & - & - & - & - & - & - & - \\
% \bottomrule
% \end{tabular}
% \caption{Generation result}
% \label{table:explicit}
% \end{table*}

%% file: text/conclusion.tex
% The emergence of LLMs has opened new avenues in the embodied agent domain. This paper introduced the \textbf{\acro} task, a pioneering approach aimed at propelling embodied agents into challenging outdoor settings. 
% Further, we introduced a novel, general, mechanism for using LLMs to reason about robot plans in unseen environments, 
% %We employed the Rapidly exploring random tree (RRT) methodology as a premier solution to this task. 
% %Furthermore, we 
% and 
% proposed the \textbf{CASR}, the first metric to assess balance between reasoning and action for embodied agents. 
% %As we move forward, these innovations offer promising directions in harnessing the potential of LLMs for embodied agent advancement
% Our formulation more closely mirrors how humans navigate and explore, trading off between thinking and acting to both leverage what we know in general and can see in the specific.  
% \vspace{-0.5em}
We present Embodied-RAG, a nonparametric embodied memory system capable of capturing embodied memories at any spatial and semantic resolution in both indoor and outdoor environments, and retrieving and generating responses for navigation and explanation requests. Additionally, we introduce the Embodied-Experiences datasets for allowing the community to continue testing different RAG system for robotics settings. Our findings demonstrate that Embodied-RAG can outperform existing baselines in all explicit, implicit, and global quires, while able to build the structured graph-memory 9.76 times faster than LightRAG. Our results indicate that Embodied-RAG shows potential as the basis for incorporating large nonparameteric embodied memories into foundation models.  The memories constructed here are open-world and semantically rich while still being tied to the environment. This provides a fundamentally new resource to robotic systems and we are excited for future extensions to manipulation and dynamic environments that enable robotics tasks out of reach for current approaches.
%as a task-, environment-, and platform-agnostic pipeline for general-purpose robots that can operate anywhere humans exist and care about. 

%% file: root.bbl
\begin{thebibliography}{10}
\providecommand{\url}[1]{#1}
\csname url@samestyle\endcsname
\providecommand{\newblock}{\relax}
\providecommand{\bibinfo}[2]{#2}
\providecommand{\BIBentrySTDinterwordspacing}{\spaceskip=0pt\relax}
\providecommand{\BIBentryALTinterwordstretchfactor}{4}
\providecommand{\BIBentryALTinterwordspacing}{\spaceskip=\fontdimen2\font plus
\BIBentryALTinterwordstretchfactor\fontdimen3\font minus \fontdimen4\font\relax}
\providecommand{\BIBforeignlanguage}[2]{{%
\expandafter\ifx\csname l@#1\endcsname\relax
\typeout{** WARNING: IEEEtran.bst: No hyphenation pattern has been}%
\typeout{** loaded for the language `#1'. Using the pattern for}%
\typeout{** the default language instead.}%
\else
\language=\csname l@#1\endcsname
\fi
#2}}
\providecommand{\BIBdecl}{\relax}
\BIBdecl

\bibitem{rag1}
P.~Lewis, D.~Kiela \emph{et~al.}, ``Retrieval-augmented generation for knowledge-intensive nlp tasks,'' 2021.

\bibitem{rag2}
A.~Asai, S.~Min, Z.~Zhong, and D.~Chen, ``Acl 2023 tutorial: Retrieval-based language models and applications,'' \emph{ACL 2023}, 2023.

\bibitem{rag3}
J.~Chen, H.~Lin, X.~Han, and L.~Sun, ``Benchmarking large language models in retrieval-augmented generation,'' 2023.

\bibitem{Hughes2022HydraAR}
N.~Hughes \emph{et~al.}, ``Hydra: A real-time spatial perception system for 3d scene graph construction and optimization,'' \emph{RSS}, 2022.

\bibitem{graphrag}
D.~Edge, H.~Trinh, N.~Cheng, J.~Bradley, A.~Chao, A.~Mody, S.~Truitt, and J.~Larson, ``From local to global: A graph rag approach to query-focused summarization,'' \emph{arXiv preprint arXiv:2404.16130}, 2024.

\bibitem{lightrag}
\BIBentryALTinterwordspacing
Z.~Guo, L.~Xia, Y.~Yu, T.~Ao, and C.~Huang, ``Lightrag: Simple and fast retrieval-augmented generation,'' 2024. [Online]. Available: \url{https://arxiv.org/abs/2410.05779}
\BIBentrySTDinterwordspacing

\bibitem{ram2023}
\BIBentryALTinterwordspacing
O.~Ram, Y.~Levine, I.~Dalmedigos, D.~Muhlgay, A.~Shashua, K.~Leyton-Brown, and Y.~Shoham, ``In-context retrieval-augmented language models,'' 2023. [Online]. Available: \url{https://arxiv.org/abs/2302.00083}
\BIBentrySTDinterwordspacing

\bibitem{fan2024surveyragmeetingllms}
\BIBentryALTinterwordspacing
W.~Fan, Y.~Ding, L.~Ning, S.~Wang, H.~Li, D.~Yin, T.-S. Chua, and Q.~Li, ``A survey on rag meeting llms: Towards retrieval-augmented large language models,'' 2024. [Online]. Available: \url{https://arxiv.org/abs/2405.06211}
\BIBentrySTDinterwordspacing

\bibitem{gao2024rag}
\BIBentryALTinterwordspacing
Y.~Gao, Y.~Xiong, X.~Gao, K.~Jia, J.~Pan, Y.~Bi, Y.~Dai, J.~Sun, M.~Wang, and H.~Wang, ``Retrieval-augmented generation for large language models: A survey,'' 2024. [Online]. Available: \url{https://arxiv.org/abs/2312.10997}
\BIBentrySTDinterwordspacing

\bibitem{gao2022precisezeroshotdenseretrieval}
\BIBentryALTinterwordspacing
L.~Gao, X.~Ma, J.~Lin, and J.~Callan, ``Precise zero-shot dense retrieval without relevance labels,'' 2022. [Online]. Available: \url{https://arxiv.org/abs/2212.10496}
\BIBentrySTDinterwordspacing

\bibitem{chan2024rqraglearningrefinequeries}
\BIBentryALTinterwordspacing
C.-M. Chan, C.~Xu, R.~Yuan, H.~Luo, W.~Xue, Y.~Guo, and J.~Fu, ``Rq-rag: Learning to refine queries for retrieval augmented generation,'' 2024. [Online]. Available: \url{https://arxiv.org/abs/2404.00610}
\BIBentrySTDinterwordspacing

\bibitem{min2021film}
S.~Y. Min, D.~S. Chaplot, P.~Ravikumar, Y.~Bisk, and R.~Salakhutdinov, ``Film: Following instructions in language with modular methods,'' \emph{ICLR}, 2021.

\bibitem{min2022don}
S.~Y. Min, Yonatan \emph{et~al.}, ``Don't copy the teacher: Data and model challenges in embodied dialogue,'' \emph{EMNLP}, 2022.

\bibitem{chaplot2020object}
Chaplot, R.~R \emph{et~al.}, ``Object goal navigation using goal-oriented semantic exploration,'' \emph{NeurIPS}, vol.~33, 2020.

\bibitem{shafiullah2022clipfields}
N.~M.~M. Shafiullah, C.~Paxton, L.~Pinto, S.~Chintala, and A.~Szlam, ``Clip-fields: Weakly supervised semantic fields for robotic memory,'' \emph{arXiv: Arxiv-2210.05663}, 2022.

\bibitem{huang23vlmaps}
C.~Huang, O.~Mees, A.~Zeng, and W.~Burgard, ``Visual language maps for robot navigation,'' in \emph{Proceedings of the ICRA}, London, UK, 2023.

\bibitem{chang2023goat}
M.~Chang, T.~Gervet, M.~Khanna, S.~Yenamandra, D.~Shah, S.~Y. Min, K.~Shah, C.~Paxton, S.~Gupta, D.~Batra \emph{et~al.}, ``Goat: Go to any thing,'' \emph{arXiv:2311.06430}, 2023.

\bibitem{ramakrishnan2022poni}
S.~K. Ramakrishnan, D.~S. Chaplot, Z.~Al-Halah, J.~Malik, and K.~Grauman, ``Poni: Potential functions for objectgoal navigation with interaction-free learning,'' in \emph{ICCV}, 2022, pp. 18\,890--18\,900.

\bibitem{min2023self}
S.~Y. Min, Y.-H.~H. Tsai, W.~Ding, A.~Farhadi, R.~Salakhutdinov, Y.~Bisk, and J.~Zhang, ``Self-supervised object goal navigation with in-situ finetuning,'' in \emph{2023 IROS}.\hskip 1em plus 0.5em minus 0.4em\relax IEEE, 2023, pp. 7119--7126.

\bibitem{scenegraph}
Li, Fuchun \emph{et~al.}, ``Embodied semantic scene graph generation,'' in \emph{CoRL}, A.~Faust, D.~Hsu, and G.~Neumann, Eds.\hskip 1em plus 0.5em minus 0.4em\relax PMLR, 2022.

\bibitem{sayplan}
K.~Rana \emph{et~al.}, ``Sayplan: Grounding large language models using 3d scene graphs for scalable task planning,'' in \emph{CoRL}, 2023.

\bibitem{hornung2013octomap}
A.~Hornung, K.~M. Wurm, M.~Bennewitz, C.~Stachniss, and W.~Burgard, ``Octomap: An efficient probabilistic 3d mapping framework based on octrees,'' \emph{Autonomous robots}, vol.~34, pp. 189--206, 2013.

\bibitem{zhang2018semantic}
L.~Zhang, L.~Wei, P.~Shen, W.~Wei, G.~Zhu, and J.~Song, ``Semantic slam based on object detection and improved octomap,'' \emph{IEEE Access}, vol.~6, pp. 75\,545--75\,559, 2018.

\bibitem{zheng2023asystem}
K.~Zheng, A.~Paul, and S.~Tellex, ``Asystem for generalized 3d multi-object search,'' in \emph{2023 ICRA}.\hskip 1em plus 0.5em minus 0.4em\relax IEEE, 2023, pp. 1638--1644.

\bibitem{liu2019object}
K.~Liu, Z.~Fan, M.~Liu, and S.~Zhang, ``Object-aware semantic mapping of indoor scenes using octomap,'' in \emph{2019 Chinese Control Conference (CCC)}.\hskip 1em plus 0.5em minus 0.4em\relax IEEE, 2019, pp. 8671--8676.

\bibitem{anderson2018evaluation}
P.~Anderson, A.~Chang, D.~S. Chaplot, A.~Dosovitskiy, S.~Gupta, V.~Koltun, J.~Kosecka, J.~Malik, R.~Mottaghi, M.~Savva \emph{et~al.}, ``On evaluation of embodied navigation agents,'' \emph{arXiv preprint arXiv:1807.06757}, 2018.

\bibitem{imagenav}
L.~Mezghan, S.~Sukhbaatar, T.~Lavril, O.~Maksymets, D.~Batra, P.~Bojanowski, and K.~Alahari, ``Memory-augmented reinforcement learning for image-goal navigation,'' in \emph{2022 IROS}.\hskip 1em plus 0.5em minus 0.4em\relax IEEE, 2022, pp. 3316--3323.

\bibitem{zhu2017target}
Y.~Zhu, R.~Mottaghi, E.~Kolve, J.~J. Lim, A.~Gupta, L.~Fei-Fei, and A.~Farhadi, ``Target-driven visual navigation in indoor scenes using deep reinforcement learning,'' in \emph{2017 ICRA}.\hskip 1em plus 0.5em minus 0.4em\relax IEEE, 2017, pp. 3357--3364.

\bibitem{krantz2022instance}
J.~Krantz, S.~Lee, J.~Malik, D.~Batra, and D.~S. Chaplot, ``Instance-specific image goal navigation: Training embodied agents to find object instances,'' \emph{CVPR}, 2022.

\bibitem{vlnsurvey}
J.~Gu, E.~Stefani, Q.~Wu, J.~Thomason, and X.~E. Wang, ``Vision-and-language navigation: A survey of tasks, methods, and future directions,'' \emph{arXiv:2203.12667}, 2022.

\bibitem{khanna2024goat}
Khanna, Roozbeh \emph{et~al.}, ``Goat-bench: A benchmark for multi-modal lifelong navigation,'' \emph{arXiv:2404.06609}, 2024.

\bibitem{das2018embodied}
A.~Das, S.~Datta, G.~Gkioxari, S.~Lee, D.~Parikh, and D.~Batra, ``Embodied question answering,'' in \emph{CVPR}, 2018, pp. 1--10.

\bibitem{padalkar2023open}
Padalkar \emph{et~al.}, ``Open x-embodiment: Robotic learning datasets and rt-x models,'' \emph{arXiv:2310.08864}, 2023.

\bibitem{yu2019multi}
L.~Yu, X.~Chen, G.~Gkioxari, M.~Bansal, T.~L. Berg, and D.~Batra, ``Multi-target embodied question answering,'' in \emph{ICCV}, 2019, p. 6309.

\bibitem{tan2023knowledge}
S.~Tan, M.~Ge, D.~Guo, H.~Liu, and F.~Sun, ``Knowledge-based embodied question answering,'' \emph{IEEE Transactions on Pattern Analysis and Machine Intelligence}, 2023.

\bibitem{zhong2022video}
Zhong, Tat-Seng \emph{et~al.}, ``Video question answering: Datasets, algorithms and challenges,'' \emph{arXiv:2203.01225}, 2022.

\bibitem{yang2003videoqa}
H.~Yang, L.~Chaisorn, Y.~Zhao, S.-Y. Neo, and T.-S. Chua, ``Videoqa: question answering on news video,'' in \emph{Proceedings of the eleventh ACM international conference on Multimedia}, 2003, pp. 632--641.

\bibitem{castro2020lifeqa}
Castro, Rada \emph{et~al.}, ``Lifeqa: A real-life dataset for video question answering,'' in \emph{LREC}, 2020.

\bibitem{xiao2021next}
J.~Xiao, X.~Shang, A.~Yao, and T.-S. Chua, ``Next-qa: Next phase of question-answering to explaining temporal actions,'' in \emph{ICCV}, 2021.

\bibitem{sneath1973numerical}
P.~H. Sneath and R.~R. Sokal, \emph{Numerical Taxonomy: The Principles and Practice of Numerical Classification}.\hskip 1em plus 0.5em minus 0.4em\relax W.H. Freeman, 1973.

\bibitem{Mllner2011ModernHA}
\BIBentryALTinterwordspacing
D.~M{\"u}llner, ``Modern hierarchical, agglomerative clustering algorithms,'' \emph{ArXiv}, vol. abs/1109.2378, 2011. [Online]. Available: \url{https://api.semanticscholar.org/CorpusID:8490224}
\BIBentrySTDinterwordspacing

\bibitem{yao2024tree}
S.~Yao, D.~Yu, J.~Zhao, I.~Shafran, T.~Griffiths, Y.~Cao, and K.~Narasimhan, ``Tree of thoughts: Deliberate problem solving with large language models,'' \emph{NeurIPS}, vol.~36, 2024.

\bibitem{shah2018airsim}
S.~Shah, D.~Dey, C.~Lovett, and A.~Kapoor, ``Airsim: High-fidelity visual and physical simulation for autonomous vehicles,'' in \emph{FSR}, 2018.

\end{thebibliography}
